\documentclass[sigconf]{acmart}

\AtBeginDocument{%
  \providecommand\BibTeX{{%
    \normalfont B\kern-0.5em{\scshape i\kern-0.25em b}\kern-0.8em\TeX}}}

\copyrightyear{2023}
\acmYear{2023}
\setcopyright{acmlicensed}
\acmConference[CIKM '23]{Proceedings of the 32nd ACM International Conference on Information and Knowledge Management}{October 21--25, 2023}{Birmingham, United Kingdom}
\acmBooktitle{Proceedings of the 32nd ACM International Conference on Information and Knowledge Management (CIKM '23), October 21--25, 2023, Birmingham, United Kingdom}
\acmPrice{15.00}
\acmDOI{10.1145/3583780.3614946}
\acmISBN{979-8-4007-0124-5/23/10}

\usepackage{amsmath,utfsym,wasysym,subfigure,enumitem,multirow}
\begin{document}

\title{Knowledge-inspired Subdomain Adaptation for Cross-Domain Knowledge Transfer}

\author{Liyue Chen}
\affiliation{%
  \institution{Key Lab of High Confidence Software Technologies (Peking University), Ministry of Education \& \\
  School of Computer Science, \\ Peking University}
  \state{Beijing}
  \country{China}
}

\author{Linian Wang}
\affiliation{%
  \institution{Key Lab of High Confidence Software Technologies (Peking University), Ministry of Education \& \\
  School of Computer Science, \\ Peking University}
  \state{Beijing}
  \country{China}}

\author{Jinyu Xu}
\author{Shuai Chen}
\authornotemark[1]
\affiliation{%
  \institution{Alipay (Hangzhou) Information \& Technology Co., Ltd}
  \city{Hangzhou}
  \country{China}}

\author{Weiqiang Wang}
\author{Wenbiao Zhao}
\affiliation{%
  \institution{Alipay (Hangzhou) Information \& Technology Co., Ltd}
  \city{Hangzhou}
  \country{China}}

\author{Qiyu Li}
\affiliation{%
   \institution{School of Electronics Engineering and Computer Science, Peking University}
\city{Beijing}
\country{China}
}

\author{Leye Wang}
\affiliation{%
  \institution{Key Lab of High Confidence Software Technologies (Peking University), Ministry of Education \& \\
  School of Computer Science, \\ Peking University}
  \state{Beijing}
  \country{China}
}
\authornote{Corresponding authors.}

\renewcommand{\shortauthors}{Liyue Chen et al.}

\begin{abstract}
Most state-of-the-art deep domain adaptation techniques align source and target samples in a global fashion. That is, after alignment, each source sample is expected to become similar to any target sample. However, global alignment may not always be optimal or necessary in practice. 
For example, consider cross-domain fraud detection, where there are two types of transactions: credit and non-credit. Aligning credit and non-credit transactions separately may yield better performance than global alignment, as credit transactions are unlikely to exhibit patterns similar to non-credit transactions. 
To enable such fine-grained domain adaption, we propose a novel Knowledge-Inspired Subdomain Adaptation (\textit{KISA}) framework. 
In particular, (1) We provide the theoretical insight that \textit{KISA} minimizes the shared expected loss which is the premise for the success of domain adaptation methods. (2) We propose the knowledge-inspired subdomain division problem that plays a crucial role in fine-grained domain adaption. (3) We design a knowledge fusion network to exploit diverse domain knowledge. Extensive experiments demonstrate that \textit{KISA} achieves remarkable results on fraud detection and traffic demand prediction tasks.
\end{abstract}


\maketitle

\section{Introduction} \label{section1}
Deep networks have considerably advanced the state of the art for a wide variety of real-world problems~\cite{dou2020enhancing,hete_graph_2021,bright_2022,li_dual_2022,explainable_fraud_2022}. However, we may still suffer from data scarcity when building applications to new domains (e.g., expanding business to new countries or new markets). In recent decades, we have witnessed lots of effort focusing on Unsupervised Domain Adaptation (UDA) and Semi-supervised
domain Adaptation (SSDA). This paper mainly focuses on the SSDA setting where a few
target labels are available, which becomes an important practical issue such as traffic prediction~\cite{regionTrans2019,metaST_2019,crosstres_2022}, fraud detection~\cite{zhu2020modeling,fraud_noisy_2021}, and image classification~\cite{xie2018learning,pei2018multi,zhu2020deepsubdomain}.

In recent years, there have been significant efforts to design methods for domain adaptation. These can be divided into two major categories: \textit{global domain adaptation}~\cite{ghifary2014domain,tzeng2015simultaneous,ganin2016domain,ganin2015unsupervised,pmlr_v37_long15, Sun2016DeepCC} and \textit{categorical subdomain adaptation}~\cite{long2017conditional,kumar2018co,pei2018multi,wang2018stratified,xie2018learning,zhu2020deepsubdomain}. 

Global domain adaptation methods primarily focus on aligning the global distributions between source and target domains. However, after alignment, each source sample is expected to become similar to any target sample resulting in inadequate performance. For instance, in a binary classification task, positive samples from the source domain may align with negative samples from the target domain. Therefore, even if the distribution discrepancy between source and target domains is minimized, it remains challenging to classify different categories that are close to each other. To address this issue, categorical subdomain adaptation methods \cite{no_more_dis_2017,long2017conditional} have been proposed (also known as semantic alignments \cite{xie2018learning} or conditional distribution matching \cite{long2017conditional,kumar2018co}). These methods take category information (i.e., class) into account and align conditional distributions between source and target domains. 

\begin{figure}[htbp]
  \centering
  \includegraphics[width=1\linewidth]{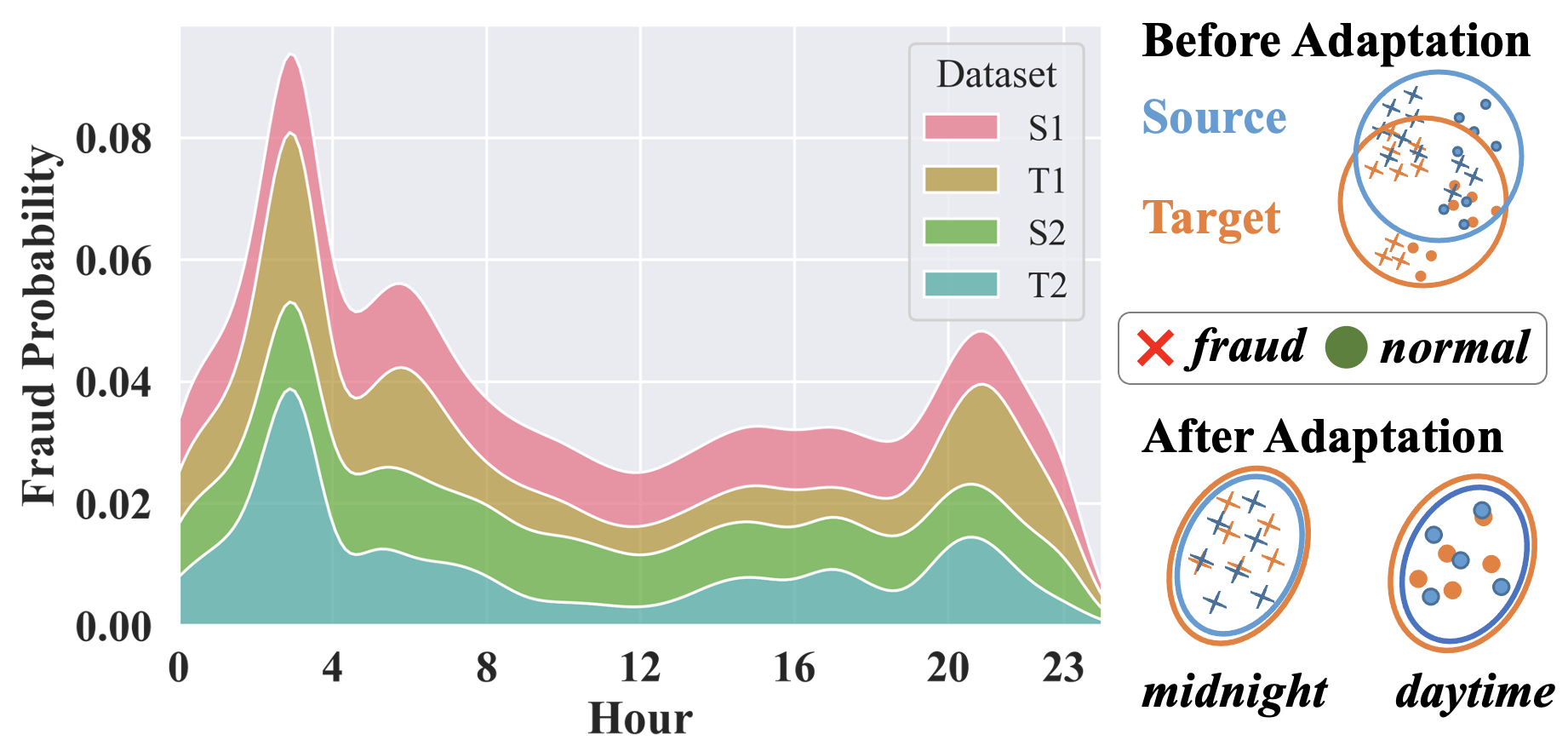}
  \vspace{-2em}
  \caption{The fraud probability at different periods. Midnight transactions are more like frauds. The analysis is on the cross-border fraud detection dataset (Sec. \ref{sec: fraud detection task}). Midnight and daytime transactions can be divided into different subdomains for alignment.}
  \vspace{-0.8em}
  \label{know_exploration}
\end{figure}

While categorical subdomain adaptation methods have proven effective in many fields \cite{long2017conditional,xie2018learning,zhu2020modeling}, they may not always be optimal as they hardly conduct fine-grained knowledge transfer. For instance, the findings depicted in Fig. ~\ref{know_exploration}, derived from an analysis conducted on the cross-border fraud detection dataset (detailed in Sec. ~\ref{sec: fraud detection task}), demonstrate midnight transactions are more like frauds. This preference among fraudsters for conducting illicit actions during this time arises from the fact that potential victims are typically asleep, rendering them less likely to notice discrepancies in their accounts \cite{cheng_st_fraud}. This observation highlights shared characteristics, such as occurrence time, among samples, leading to similar patterns in fraud probability. This phenomenon is commonly recognized as domain knowledge. 

Consequently, there is a strong motivation to utilize domain knowledge to find out similar samples and group them into subdomains that exhibit analogous patterns, enabling knowledge transfer from these aligning relevant subdomains. As highlighted by prior literature, integrating domain knowledge into domain adaptation methods can reduce uncertainty caused by limited data and enhance their effectiveness \cite{DENG2020101656}. However, designing such a knowledge-inspired subdomain adaptation framework poses several challenges:

\textbf{Challenge 1}. \textit{How to construct subdomains based on samples' similarity?} 
A subdomain should contain samples with similar properties. Both original features and their latent representation extracted by the deep networks can reflect sample similarity. A straightforward way is to concatenate original and latent representations into a new feature vector, followed by clustering to obtain subdomains. However, latent representations are more informative than the original features, and this straightforward method may be affected by poor-quality original features resulting in degraded performance. 

\textbf{Challenge 2}. \textit{How to fully utilize diverse domain knowledge?} 
There may exist multiple types of domain knowledge that may benefit an application. 
For example, in traffic prediction tasks, we may observe that (1) regions with similar functionalities (e.g., business areas) and (2) adjacent regions may exhibit comparable daily patterns \cite{yuan_functions_2012,wang2021exploring}.
Hence, it is crucial to develop a technique that can effectively incorporate different types of domain knowledge.

To address these challenges, we propose a framework called \textit{KISA}. Our main contributions include:
\begin{itemize}
    \item As far as we know, this is one of the pioneering efforts toward knowledge-inspired deep subdomain adaptation methods. Compared to global domain adaptation or categorical subdomain adaptation, our method facilitates a more fine-grained transfer learning strategy through domain knowledge.
    
    \item Specifically, \textit{KISA} proposes the knowledge-inspired subdomain division problem to construct subdomains, which is crucial for fine-grained knowledge transfer. Moreover, \textit{KISA} introduces a knowledge fusion network to fully exploit diverse domain knowledge.
    
    \item We conducted extensive experiments on cross-domain fraud detection and traffic demand prediction tasks. For each task, we explored and utilized two types of domain knowledge to facilitate fine-grained knowledge transfer. The experimental results demonstrate the effectiveness of \textit{KISA}. Compared to state-of-the-art global adaptation and categorical subdomain adaptation methods, \textit{KISA} can improve the prediction performance by up to 4.79\% and 3.17\% in fraud detection and traffic demand prediction tasks, respectively.
\end{itemize}

\section{Formulation} \label{formulation}

\noindent \textbf{Definition 1. Semi-supervised Domain Adaptation.} Given a source domain $\mathcal{D}^{src}=\{(x_i^{src},y_i^{src})\}$ with $N^{src}$ labeled samples and a target domain $\mathcal{D}^{tgt}=\{(x_i^{tgt},y_i^{tgt})\}$ with $N^{tgt}_{train}$ labeled samples and $N^{tgt}_{test}$ unlabeled samples. Note that $N^{tgt}_{train} \ll N^{src}$ and it is called semi-supervised transfer learning problem \cite{tzeng2015simultaneous, zhu2020modeling}. The source and target domain are sampled from joint distributions $P(\textbf{x}^{src},\textbf{y}^{src})$ and $Q(\textbf{x}^{tgt},\textbf{y}^{tgt})$ respectively. $y_i$ indicates the label of $x_i$. The task is to improve the prediction performance of the unlabeled test set in the target domain with the help of $\mathcal{D}^{src}$ by optimizing a deep network $f(\textbf{x})=\textbf{y}$. 

\noindent \textbf{Definition 2. Domain Knowledge.} In this study, domain knowledge $k \in \mathcal{K}$ ($\vert\mathcal{K}\vert=K$) can be utilized to select or generate informative features $x_{p}$, satisfying the conditional entropy $H(P(\textbf{y}\vert \textbf{x} \backslash \{x_p\}))- H(P(\textbf{y}\vert \textbf{x})) > \delta$. $\delta$ is a positive number that filters less beneficial features ($\textbf{x}$ is the whole features sets). 

\begin{figure}[htbp]
  \centering
  \includegraphics[width=\linewidth]{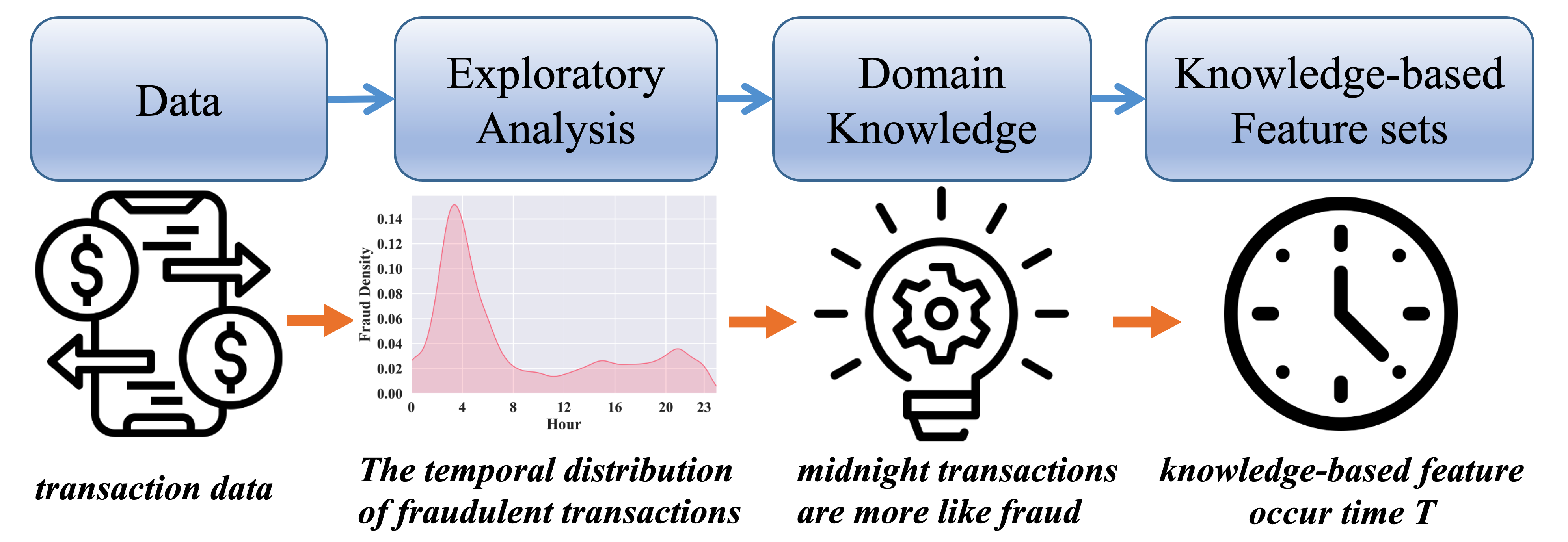}
  \vspace{-2.5em}
  \caption{Generating domain knowledge-based feature sets.}
  \vspace{-1.5em}
  \label{fig:knowledge_flow}
\end{figure}

This definition says that the prediction uncertainty is smaller by taking the features from domain knowledge. Domain knowledge is typically obtained through exploratory data analysis. For example, in Fig. \ref{fig:knowledge_flow}, we explore the fraud probability distribution at different periods and find that midnight transactions are more likely to be fraudulent than those in the daytime. This observation corresponds with previous findings \cite{cheng_st_fraud}. Note that in existing studies \cite{zhu2020modeling,cheng_fraud_2020,intention_fraud_detection_2021}, domain knowledge is often used in encoding original data into features. In this paper, we try to further incorporate domain knowledge into subdomain construction and alignment for domain adaptation.

\noindent \textbf{Definition 3. Knowledge-based Feature Sets.} 
Domain knowledge helps reduce the prediction uncertainty by adding informative features $x_p$ into $\textbf{x}\backslash \{x_p\}$ and thus the conditional entropy $H(P(\textbf{y}\vert \textbf{x}))$ and $H(P(\textbf{y}\vert x_p))$ both are small. Since domain knowledge is generalizable, in the target domain, we may still have $H(Q(\textbf{y}\vert \textbf{x} \backslash \{x_p\}))- H(Q(\textbf{y}\vert \textbf{x})) > \delta$ as well as small $H(Q(\textbf{y}\vert x_p))$. 
Therefore, $x_p$ may provide a good perspective to transfer relevant domain knowledge from source domains. 
Formally, derived from domain knowledge $k_i \in \mathcal{K}$, $x_p^i$ is further defined as the $i^{th}$ element in the knowledge-based feature sets $\mathcal{U}=\{x_p^1,...,x_p^K\}$. 

As shown in Fig. \ref{fig:knowledge_flow}, recalling that midnight transactions are more likely to be fraudulent, we could regard the transaction occurrence time as the element in the knowledge-based feature sets.

\section{Method}
\subsection{Framework Overview}
\begin{figure*}[htbp]
  \centering
  \includegraphics[width=0.98\linewidth]{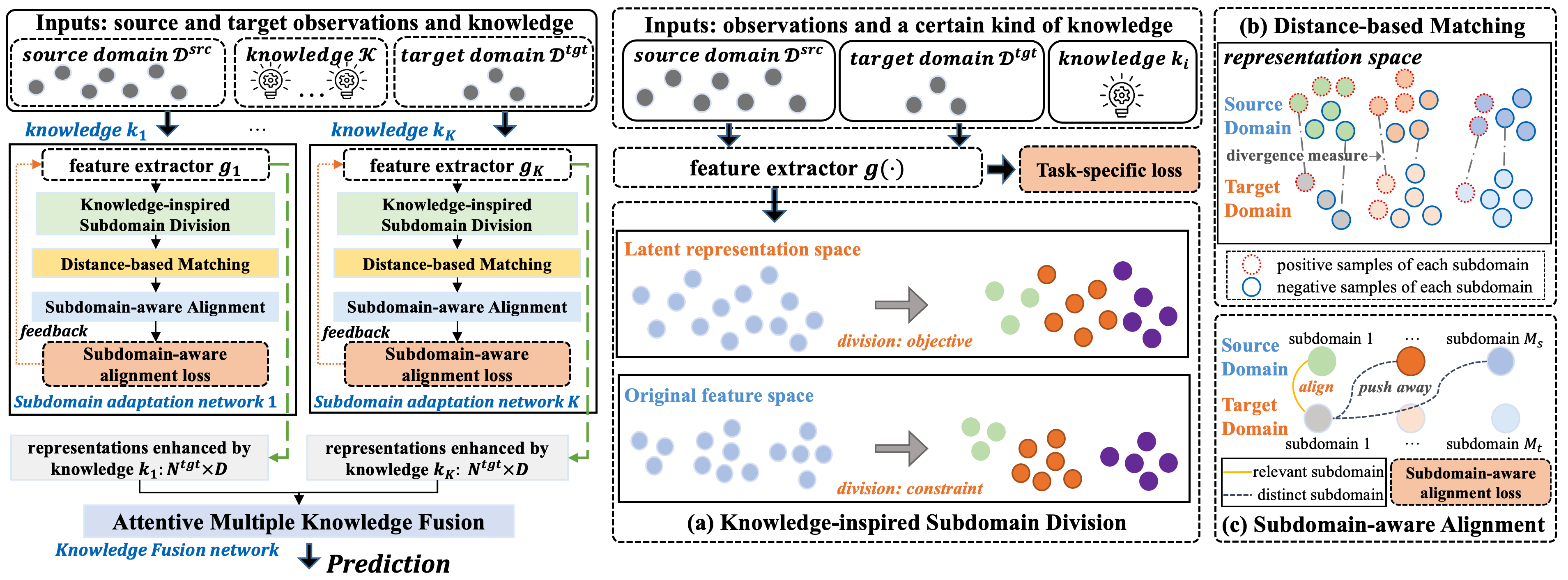}
  \vspace{-1.2em}
  \caption{Left: The structure of the \textit{KISA} framework consists of several knowledge-inspired subdomain adaptation networks and a knowledge fusion network. Right: The structure of subdomain adaptation network which consists of three components (a) knowledge-inspired subdomain division, (b) distance-based matching, and (c) subdomain-aware alignment.}
  \label{fig: framework_network}
  \vspace{-1em}
\end{figure*}

The proposed framework comprises several knowledge-inspired subdomain adaptation networks and a knowledge fusion network (Fig. \ref{fig: framework_network}). Each subdomain adaption network incorporates a specific type of domain knowledge and learns a kind of representation, which is further integrated by the knowledge fusion network. Each knowledge-inspired subdomain adaptation network consists of three cascaded components: (a) Knowledge-Inspired Subdomain Division (\textit{KISD}), (b) Distance-Based Matching (\textit{DBM}), and (c) Subdomain-Aware Alignment (\textit{SAA}). In \textit{KISD}, we use domain knowledge to construct subdomains by solving the proposed knowledge-inspired subdomain division problem. \textit{DBM} measures the distance between every subdomain and establishes matched relationships between source and target subdomains. \textit{SAA} then transfers relevant knowledge from the source domain by aligning corresponding subdomains.

\subsection{Knowledge-inspired Subdomain Division} \label{division}
\subsubsection{Problem of Knowledge-inspired Subdomain Division}
As introduced in Sec. \ref{section1}, grouping similar samples into subdomains and aligning these subdomains across domains may enhance performance. Thus, the central task is identifying shared sample properties. This can be approached through two key aspects:

\begin{itemize}[leftmargin=0.24cm]
    \item \textit{Knowledge-based features}: Samples with the close original features may be similar (e.g., for knowledge-based features $x_p$, $H(Q(\textbf{y}\vert x_p))$ is small). In spatiotemporal prediction tasks, social media check-ins can be a useful proxy \cite{yang2016participatory} and \textit{RegionTrans} \cite{regionTrans2019} further align the regions across different cities with similar check-ins patterns. Similarly, knowledge identified by literature \cite{cheng_st_fraud} (e.g., transaction time, card type) may also benefit fraud detection.
    
    \item \textit{Latent representation}: Unlabeled data can significantly impact classifier boundaries by guiding them towards regions with a low density of data points \cite{chapelle05b_pmlrvR5}. In semi-supervised learning, minimizing conditional entropy can shift predictor boundaries away from high-density areas \cite{NIPS2004_96f2b50b,miyato2018virtual}. The shared feature extractor $g(\cdot)$ has the capability to map similar samples into comparable representations. Therefore, the distance between samples in the representation space could be utilized as a similarity function.
\end{itemize}

As introduced in Sec.~\ref{section1}, concatenating the original knowledge-based features and latent representation into a new feature vector may be affected by poor-quality original features resulting in degraded clustering performance.
A more reasonable approach would be to emphasize the similarity of latent representation and constrain clustering results with similar original knowledge-based features. With this insight, we define the problem of knowledge-inspired subdomain division:

Given a set of representation points $\mathcal{Z} = \{z_i^{o}| 1\le i \le N^o\}$ ($z_i^{o}$ is the representation extracted by $g(\cdot)$, $o\in \{src,tgt\}$), cluster number $M$, and knowledge-based features $x_p^i \in \mathbb{R}^{N^o \times F}$ ($F$ is the dimension of knowledge-based features). We would like to find an optimal division $\{\mathcal{Z}_1,...,\mathcal{Z}_M\}$ by minimizing the following objectives:
\begin{align} 
    \min &\sum^{M}_{j=1}\sum_{b\in \mathcal{Z}_j} dist_r(b,\overline{\mathcal{Z}}_j) \label{clustering_object} \\
    s.t.\ \sum_{c^{\prime} \in x_{p(i)}} &dist_o(c^{\prime}, \overline{x_p}_{(i)}) \le \sum_{c \in x_{p(j)}} dist_o(c, \overline{x_p}_{(i)}) \ \ \forall i\neq j \label{clustering_constraint}
\end{align}
where $\overline{\mathcal{Z}}_{i}$ is the representation centroid of subdomain $\mathcal{Z}_i$. $x_{p(i)}$ contains the knowledge-based features belonging to subdomain $\mathcal{Z}_i$ while $\overline{x_p}_{(i)}$ is its centroid. $dist_o(\cdot)$ and $dist_r(\cdot)$ measure the distance (e.g., Euclidean distance) in the original features and latent representation space, respectively. Eq. \ref{clustering_object} aims to minimize the distance between each representation point and its corresponding representation centroid within each subdomain. Additionally, Eq. \ref{clustering_constraint} guarantees that the knowledge-based features of samples within a divided subdomain are more similar to each other than to those in other subdomains. By solving the above optimization problem, we divide $\mathcal{D}^{src}$ and $\mathcal{D}^{tgt}$ into $\mathcal{D}^{src}_{(m)}$ ($1 \le m\le M_s$), and $\mathcal{D}^{tgt}_{(n)}$ ($1 \le n\le M_t$), respectively.

\subsubsection{Solution for Knowledge-inspired Subdomain Division Problem} \label{solution_1d}
We introduce a dynamic programming \cite{ckmeans1d} solution and a community detection solution for the subdomain division problem using \textit{1-D} or high-dimensional knowledge-based features, respectively.

\textbf{\textit{Optimal division for \textit{1-D} knowledge-based features}}. We first sort the original knowledge features $x_p \in \mathbb{R}^{N \times 1}$ to ensure Eq.~\ref{clustering_constraint} is satisfied. Let $\{z_1, . . . , z_{N}\}$ be the sorted representation array extracted by the feature extractor $g(\cdot)$. Recall that Eq.~\ref{clustering_object} aims to assign elements of the sorted representation array into $M$ clusters so that the sum of squares of intra-cluster distances from each element to its corresponding centroid is minimized. To this end, we define a sub-problem as finding the minimum intra-cluster distance of clustering $z_1, ..., z_i$ into $m$ clusters. The corresponding minimum intra-cluster distance is recorded in $C[i, m]$. Let $j$ be the index of the smallest number in cluster $m$ in an optimal solution to $C[i, m]$. $C[j-1, m-1]$ must be the optimal intra-cluster distance for the first $j - 1$ points in $m-1$ clusters, for otherwise, one would have a better solution to $C[i, m]$. This establishes the optimal substructure for dynamic programming and leads to the recurrence equation:
\begin{equation}
    C[i, m] = \min_{m\le j \le i} \{ C[j-1,m-1] + d(z_j,...,z_i) \}
\end{equation}
where $d(z_j,....,z_i)$ is the sum of squared distances from $z_j,...,z_i$ to their centroid. The above process is initialized with $C[0,0] = 0$. This algorithm requires $O(N^2M)$ time to iteratively compute $d(z_j, . . . , z_i)$ using a recurrence structure. However, in practice, when $N$ is very large, this algorithm may take too much time to compute. To address this issue, we can predefine $B$ split points ($B \ll N$), and the dynamic program will only split the samples among these preset split points. This reduces the time complexity to $O(B^2M)$.

\textbf{\textit{Division for high-dimension knowledge-based features}}. We first construct the graph $G=(V, E)$, where every sample is a node in the graph. The edges between two nodes $i$ and $j$ are determined using the following equation:
\begin{equation}
e(i, j) = 
\begin{cases} \phantom{-} \frac{1}{dist_r(z_i,z_j)} & \text{if } dist_o(x_p^i,x_p^j) \le \kappa \\ 
\ \ \ \ \ \ \ \ 0 & \textit{, otherwise}
\end{cases}
\end{equation}
where $\kappa$ is the threshold that controls the number of edges. Samples closer than $\kappa$ in the original feature space connect via edges. We apply community detection algorithms such as label propagation \cite{lp_2002learning} to identify subdomains within the graph. Samples with close latent representations have larger edge weights, prompting them to be grouped into the same subdomain (with identical labels). The threshold value $\kappa$ filters out edges between samples that are far apart in the original feature space, ensuring that Eq.~\ref{clustering_constraint} remains valid.

\subsection{Distance-based Matching}\label{matching}
Getting several subdomains in the source and target domain, we then introduce a distance-based matching module to help the target subdomain utilize relevant source subdomains. By minimizing loss, we move predictor boundaries away from high-density regions. This allows us to treat the distance between samples in representation space as a similarity metric~\cite{NIPS2004_96f2b50b,miyato2018virtual}. Several distance metrics have been extensively studied to measure the discrepancy between source and target domains. These include Maximum Mean Discrepancy (MMD) \cite{pmlr_v37_long15,long2017deep}, Central Moment Discrepancy (CMD) \cite{zellinger2017central}, second-order statistics \cite{Sun2016DeepCC}), and reverse Kullback-Leibler divergence \cite{nguyen2022kl}. In this paper, we opt for MMD~\cite{gretton2012kernel} as our subdomain divergence measure due to its quick computation using the kernel function. Formally, the divergence of $i^{th}$ source subdomain and the $j^{th}$ target subdomain is defined as: 
\begin{equation}
    d_\mathcal{H}(p_{(i)},q_{(j)}) \triangleq \mathbf{E}_c \Vert \textbf{E}_{p^c_{(i)}}[\phi(z^{src}_{(i)})] - \textbf{E}_{q^c_{(j)}}[\phi(z^{tgt}_{(j)})] \Vert ^2_{\mathcal{H}}
\end{equation}
where $\mathcal{H}$ is the reproducing kernel Hilbert space (RKHS) and $\phi(\cdot)$ is the feature transformation that maps the original samples to RKHS. $\mathbf{E}_c[\cdot]$ is the mathematical expectation of the class (i.e., label). $p^c_{(m)}$ and $q^c_{(n)}$ are the distributions of class $c$ in $\mathcal{D}_{(m)}^{src}$ and $\mathcal{D}_{(n)}^{tgt}$, respectively. Since the divergence is calculated individually for each class, samples in the target subdomain will seek samples with the same label from the source subdomain to calculate their discrepancy. Then the similarity function $\mathcal{S}(\cdot,\cdot)$ is further formulated as: 
\begin{equation}
    \mathcal{S}(\mathcal{D}^{src}_{(m)},\mathcal{D}^{tgt}_{(n)}) = \frac{1}{d_\mathcal{H}(p_{(m)},q_{(n)})} 
\end{equation}
Based on the similarity function, the target subdomain may further select one or several most relevant source subdomains with the highest similarity score. We record the matching relationship (0 or 1) in matrix $R \in \mathbb{R}^{M_s\times M_t}$ ($R_{i,j}=1$ represents the $i^{th}$ source subdomain and the $j^{th}$ target subdomain are relevant).

\subsection{Subdomain-aware Alignment}
In many real-world problems, domain distribution may be seriously unbalanced (e.g., in fraud detection and early sepsis prediction applications, the number of non-fraud or healthy samples is much larger than the abnormal samples~\cite{behavior_pretrain_2022,auc_fraud_2022,early_sepsis_2023}). Hence, merely aligning the marginal and conditional distributions (also known as intra-class discrepancy) would lead to unsatisfying performance. To address this issue, borrowing ideas from Zhu et al. \cite{zhu2020modeling}, we extend the class-aware discrepancy to the subdomain-aware discrepancy, which explicitly takes the subdomain information into account and measures the intra-subdomain and inter-subdomain discrepancy across domains. The subdomain-aware alignment loss is:
\begin{equation} \label{sd_aware_loss}
    \hat{L}_{SAL}(p,q) = \sum_{j=1}^{M_t} \frac{\sum_{i=1}^{M_s} R_{i,j} \cdot d_\mathcal{H}(p_{(i)},q_{(j)})}{\sum_{i=1}^{M_s} (1-R_{i,j}) \cdot d_\mathcal{H}(p_{(i)},q_{(j)})}
\end{equation}
where $M_s$ and $M_t$ denote the number of source and target subdomains, respectively. $R_{i,j}\in \{0,1\}$ record the matching relationship between the $i^{th}$ source subdomain and the $j^{th}$ target subdomain. 

\subsection{Attentive Multiple Knowledge Fusion} \label{fusion}
As shown in Fig. \ref{fig: framework_network}, every subdomain adaptation network learns its feature extractor, and finally, we could obtain $K$ kinds of representation for each sample. An attentive fusion mechanism \cite{Zhang_Gao_Ma_Wang_Wang_Tang_2021} is leveraged to build a comprehensive and robust representation. Specifically, a learnable weight vector ($\alpha \in \mathbb{R}^{K}$) is to determine the importance of the above knowledge-inspired representation.
\begin{equation}
    \alpha^i = \mathop{Sigmoid}(W^iz^i),\ 1\le i \le K
\end{equation}
where $W^i\in \mathbb{R}^{1\times D}$ is the projection matrices and $z^i \in \mathbb{R}^{D\times 1}$ is the representation extracted by $i^{th}$ subdomain adaptation network. To make sure the representations are comparable, we constraint $\sum_i^K \beta^i=1$ by conducting normalization $\beta^i=\frac{\alpha^i}{\sum_j \alpha^j}$. Then the final fused representation is:
\begin{equation}
    h = \sum_i^K \beta^i \cdot z^i
\end{equation}
Lastly, the output layer could give the final prediction via several fully-connected layers, where Sigmoid or Tanh activation functions are used for classification and regression tasks, respectively:
\begin{equation}
    \hat{y} = \mathop{MLPs}(h)
\end{equation}

\subsection{Network Parameter Optimization}
\subsubsection{Knowledge-inspired Subdomain Adaptation Network Optimization}
Every subdomain adaptation network aims to align the distributions of relevant subdomains inspired by one kind of domain knowledge. Combining the task-specific loss and subdomain alignment loss, the loss of the subdomain adaptation network is:
\begin{equation} 
    \min_f \mathbb{E}_{(\textbf{x},y)\in\mathcal{D}^{src},\mathcal{D}^{tgt}_{train}}J(f(\textbf{x}),y)+\lambda \hat{L}_{SAL}(p,q)
\end{equation}
where $J(\cdot, \cdot)$ is the task-specific loss (e.g., cross-entropy loss for classification tasks or mean squared error loss for regression tasks). $\hat{L}_{SAL}(\cdot, \cdot)$ is the subdomain-aware alignment loss. $\lambda > 0$ is the trade-off parameter between these two losses. The training process of each subdomain adaptation network (corresponding to each type of domain knowledge) is independent, so the training can be conducted in a parallel manner.

\subsubsection{Knowledge Fusion Network Optimization}
After training every knowledge-inspired subdomain adaptation network, we obtain a series of feature extractors $g_1(\cdot),...,g_K(\cdot)$. The knowledge fusion network aims to make full use of all kinds of representation and eventually give predictions for the target domain. The loss of the knowledge fusion network is:
\begin{equation}
    \min_{\Theta} \mathbb{E}_{(\textbf{x},y)\in\mathcal{D}^{tgt}_{train}}J(\theta(g_1(\textbf{x}),...,g_K(\textbf{x}),y)
\end{equation}
where $\Theta$ is the learnable parameters mentioned in Sec. \ref{fusion}. $\theta(\cdot)$ is the learned mapping from diverse representations to prediction. 

\subsection{Theoretical Insight}

\noindent \textbf{Theorem 1 (Ben David et al. \cite{ben2010theory}}) Let $\mathcal{H}$ be the common hypothesis class for source and target. The expected error for the target domain is upper bounded as: 
\begin{equation}
    \epsilon_t(h) \leq \epsilon_s(h) + \frac{1}{2}d_{\mathcal{H}\Delta\mathcal{H}}(P,Q) + C, \forall h \in \mathcal{H}
\end{equation}
where $\epsilon_s(h)$ is the expected error of $h$ on the source domain and $d_{\mathcal{H}\Delta\mathcal{H}}(P,Q)$ is the domain divergence measure by a discrepancy distance between two distributions. 

Many domain adaptation methods aim to align global distribution between the source domain and target domain such that $P$ and $Q$ are close. $C = \min_{h\in \mathcal{H}} R_{\mathcal{S}}(h,f_{\mathcal{S}}) + R_{\mathcal{T}}(h,f_{\mathcal{T}})$ is the shared expected loss that is expected to be negligibly small and usually disregarded by previous methods \cite{pmlr_v37_long15}. However, if $C$ is large, we cannot expect to learn a good target classifier by minimizing the source error \cite{ben2010theory, xie2018learning}.

Referring to the work of categorical subdomain adaptation \cite{xie2018learning,zhu2020deepsubdomain}, the shared expected loss $C$ can be decomposed as the following four items based on the triangle inequality for classification error \cite{crammer_JMLR,ben2010theory} which says that for any labeling functions $f_1, f_2, f_3$, we have $R(f_1,f_2) \le R(f_1,f_3) + R(f_2,f_3)$:
\begin{align}
    C &= \min_{h\in \mathcal{H}} R_{\mathcal{S}}(h,f_{\mathcal{S}}) + R_{\mathcal{T}}(h,f_{\mathcal{T}}) \\
    &\le \min_{h\in \mathcal{H}} R_{\mathcal{S}}(h,f_{\mathcal{S}}) + R_{\mathcal{T}}(h,f_{\mathcal{S}}) + R_{\mathcal{T}}(f_{\mathcal{S}},f_{\mathcal{T}}) \\
    &\le \min_{h\in \mathcal{H}} R_{\mathcal{S}}(h,f_{\mathcal{S}}) + R_{\mathcal{T}}(h,f_{\mathcal{S}}) + R_{\mathcal{T}}(f_{\mathcal{S}},f_{\mathcal{\hat{T}}}) + R_{\mathcal{T}}(f_{\mathcal{T}},f_{\mathcal{\hat{T}}}) \label{last_ineq}
\end{align}
where $f_{\mathcal{S}}$ and $f_{\mathcal{T}}$ are true labeling functions for the source and target domain, respectively. The first two terms should be small since $h$ is learned with the labeled source samples. The last term $R_{\mathcal{T}}(f_{\mathcal{T}},f_{\mathcal{\hat{T}}})$ denotes the disagreement between the ideal target labeling function $f_{\mathcal{T}}$ and the learned labeling function $f_{\mathcal{\hat{T}}}$, which would be optimized during the learning process. 

We then mainly focus the third item $R_{\mathcal{T}}(f_{\mathcal{S}},f_{\mathcal{\hat{T}}})$. Note that the hypothesis $h$ could be decomposed into the feature extractor $G$ and classifier $F$. The third item could be further rewritten as 
\begin{equation}
R_{\mathcal{T}}(f_{\mathcal{S}},f_{\mathcal{\hat{T}}}) = \mathbb{E}_{x\sim \mathcal{T}}[l(F_{\mathcal{S}}\circ G(x),F_{\mathcal{\hat{T}}} \circ G(x))] 
\label{eq:rt}
\end{equation}
where $l(\cdot)$ is typically 0-1 loss function. For clarity, we assign the index $k$ to the source subdomain that is relevant with $k^{th}$ target subdomain. After aligning them, we have $\mathbb{E}_{x\sim \mathcal{S}^{(k)}}G(x)=\mathbb{E}_{x\sim \mathcal{T}^{(k)}}G(x)$. The above item will be small if the source labeling function $F_{\mathcal{S}}$ and the learned labeling function $F_\mathcal{\hat{T}}$ give the same prediction for the target domain samples. Then, Eq.~\ref{eq:rt} is written as,
\begin{equation}
R_{\mathcal{T}}(f_{\mathcal{S}},f_{\mathcal{\hat{T}}}) = \mathbb{E}_k [\mathbb{E}_{x\sim \mathcal{T}^{(k)}}[l(F_{\mathcal{S}^{(k)}}\circ G(x),F_{\mathcal{\hat{T}}^{(k)}} \circ G(x))]] 
\label{eq:rt_2}
\end{equation}
For the samples in target subdomain $k$ (i.e., $x\sim \mathcal{T}^{(k)}$), we have $P_{(k)}(\textbf{y}\vert x_p) \approx Q_{(k)}(\textbf{y}\vert x_p)$ (recall that we align the conditional distribution and their conditional entropy both are small as introduced in Definition 3), so the classifiers $F_{\mathcal{S}^{(k)}}$ and $F_{\mathcal{\hat{T}}^{(k)}}$) will give similar predictions. Hence, for $x\sim \mathcal{T}^{(k)}$, we may infer
\begin{equation}
    \mathbb{E}_{x\sim \mathcal{T}^{(k)}} l(F_{\mathcal{S}^{(k)}}\circ G(x),F_{\mathcal{\hat{T}}^{(k)}} \circ G(x)) \le \mathbb{E}_{x\sim \mathcal{T}^{(k)}} l(F_{\mathcal{S}}\circ G(x),F_{\mathcal{\hat{T}}} \circ G(x))
    \label{eq:rt_k}
\end{equation}
if we find a good pair of matching subdomains $\{\mathcal{S}^{(k)}, \mathcal{T}^{(k)}\}$.
Consequently, the third item $R_{\mathcal{T}}(f_{\mathcal{S}},f_{\mathcal{\hat{T}}})$ is expected to be small.

\section{Experiment}
\label{sec:experiment}

\subsection{Cross-border Fraud Detection}
\label{sec: fraud detection task}
\subsubsection{Fraud Detection Dataset and Task} 
We collected four fraud detection datasets from a leading cross-border e-commerce company. S1 and S2 are from high-activity, engaged countries; T1 and T2 are from newly opened countries with data scarcity. S1 and S2 are the source domain, while T1 and T2 are the target domain. T2 has fewer samples compared to T1, with less conspicuous midnight fraud transaction patterns (see Fig.~\ref{know_exploration}). Table \ref{table:sampled_dataset_pos_neg_number} illustrates the class imbalance issue in all datasets, with significantly more negative samples (normal cases).

The datasets record users' historical event sequences and current payment event behavior. Each event note details like routermac, trade amount, and occurrence time. We partitioned the data into chronological training, validation, and test sets. The objective is binary fraud prediction for the current payment event.

\vspace{-.3em}
\begin{table}[htbp]
\caption{Statistics of the fraud detection datasets.}
\vspace{-.8em}
\resizebox{0.485\textwidth}{!}{
\begin{tabular}{lccccccccccc}
\toprule
\textbf{Dataset} & \textbf{S1} & \textbf{T1} & \textbf{S2} & \textbf{T2} \\ 
\midrule
Training period & 01.01-05.15 
& 04.15-05.15& 03.01-06.30 
& 06.01-06.30\\
Validation period & N/A 
& 05.16-05.31 
& N/A & 07.01-07.14\\
Test period & N/A & 06.01-07.01 & N/A & 07.15-08.15  \\
\hline
\# Sequences & 192.46k  & 3.80k & 180.70k & 2.92k\\
\# Events & 1,701.03k & 66.95k & 1,559.55k  & 51.65k \\
\# Fields & 96 & 96 & 96 & 96\\
\% Frauds  & 10.96\% & 8.17\% & 7.52\% & 9.08\%  \\ 
\bottomrule
\end{tabular}}
\vspace{-1em}
\label{table:sampled_dataset_pos_neg_number}   
\end{table}

\subsubsection{Baseline Models}
\begin{itemize}[leftmargin=0.24cm]
\item \textbf{\textit{LSTM}} \cite{wang_2017_pkdd,jurgovsky_sequence_2018} and \textbf{\textit{LSTM (FT)}}: are based on the LSTM (Long Short-Term Memory) model. Only target domain data are used to train \textit{LSTM} while \textit{LSTM (FT)} trains based on the source domain data and then fine-tunes using the limited target domain data. 

\item \textbf{\textit{MMD}} ~\cite{Tzeng_invariance} is a global domain adaptation method that uses MMD (Maximum Mean Discrepancy) to align the distribution between the source and target domain. Its loss function contains the MMD loss and the cross-entropy loss.

\item \textbf{\textit{Transfer-HEN}} \cite{zhu2020modeling} is a categorical subdomain adaptation method that exploits categorical information to construct subdomains (i.e., the subdomain consists of samples within the same class). It minimizes the distance between the subdomains with the same label across the source and target domain while maximizing the distance between the subdomains with different labels.

\item \textbf{\textit{DSAN}} \cite{zhu2020deepsubdomain} is a categorical subdomain adaptation method that aligns the relevant subdomain distributions across different domains based on LMMD (Local Maximum Mean Discrepancy). LMMD adds a weight coefficient to the MMD formula. \textit{DSAN} generates the pseudo label for unlabeled data, then uses the label to construct subdomains and calculate LMMD.

\item \textbf{\textit{KL}} \cite{nguyen2022kl} is a global domain adaptation method that minimizes the reverse Kullback-Leibler divergence between source and target representations for better generalization to the target domain.
\end{itemize}

\subsubsection{Evaluation Metric} The fraud detection task is a binary classification task, we evaluate with AUC (Area Under ROC) and AUPRC (Area Under the Precision-Recall Curve). The AUPRC metric is suitable for evaluating highly imbalanced and skewed datasets \cite{davis_auprc} like our fraud detection datasets. 

\subsubsection{Implementation Details} 
To ensure fair comparisons, all methods use the same backbone network consisting of two stacked LSTM layers with a hidden size of 300. The fully connected network structure includes a dropout layer (with a keep probability of 0.8) to prevent overfitting and takes in embedding vectors as input, producing a 2-dimensional output indicating whether the transaction is fraudulent. The trade-off parameter $\lambda$ is set to 0.1, and training is performed using stochastic gradient descent on shuffled mini-batches with a batch size of 32. We utilize the \textit{Adagrad} optimizer \cite{adagrad_2011} with a learning rate of $10^{-4}$ and implement an early stop mechanism that halts training after no improvement for 50 epochs.

\subsubsection{Domain-knowledge Exploration \& Exploitation} \textit{KISA} leverages two kinds of domain knowledge for the fraud detection task.

\textbf{\textit{Knowledge 1: midnight transactions are more like frauds}}. 
Fraudsters tend to carry out fraudulent transactions at midnight when victims are asleep and less likely to notice changes in their accounts \cite{cheng_st_fraud}. This trend is also observed in the S1, S2, T1, and T2 datasets as depicted in Fig.~\ref{know_exploration}. Therefore, \textit{KISA-Hour} utilizes transaction time (i.e., hour) to construct subdomains. We divide the day into four periods - midnight, morning, afternoon, and evening - based on our daily routines. Consequently, we construct four subdomains ($M$ in Eq.~\ref{clustering_object}) for both source and target domains.

\textbf{\textit{Knowledge 2: credit card transactions are more like frauds.}} Credit card fraud is more common and dangerous than non-credit card fraud due to the convenience and various incentives, such as cashback and reward points. Our analysis of the fraud detection dataset also reveals that credit card transactions are more likely to be fraudulent. Therefore, \textit{KISA-CardType} utilizes transaction card type to construct subdomains. Card type (0-1 variable) is the special case of 1-D knowledge-based feature and we build two subdomains in both source and target domains.

\begin{table*}[htbp]
\caption{Results on the fraud detection task. `±' denotes the standard deviation. ${}^{*}$ represents $p<0.05$. The best results are in bold.}
\vspace{-0.8em}
\setlength{\tabcolsep}{3.7mm}{
\begin{tabular}{lcccccc}
\toprule
& \multicolumn{2}{c}{\textbf{S1} $\rightarrow$ \textbf{T1}} & \multicolumn{2}{c}{\textbf{S2} $\rightarrow$ \textbf{T2}}\\ 
 \cmidrule(lr){2-3} \cmidrule(lr){4-5} 
& AUC & AUPRC & AUC & AUPRC \\
\midrule
\textbf{Target Only} \\
\textit{LSTM}& 0.7123±0.0162 & 0.3190±0.0442  &  0.6318±0.0063 & 0.1086±0.0087 \\
\midrule
\textbf{Source \& Target} \\
\textit{LSTM (FT)} & 0.6637±0.0417 & 0.2595±0.0464 & 0.5934±0.0152 & 0.1353±0.0047 \\
\textit{MMD}  & 0.7475±0.0127 & 0.3707±0.0249 & 0.6120±0.0064 & 0.1454±0.0045\\
\textit{KL} & 0.7498±0.0048 & 0.4140±0.0111 & 0.6269±0.0135 & 0.1521±0.0078\\
\textit{Transfer-HEN} & 0.7506±0.0146 & 0.3733±0.0544 & 0.6132±0.0110 & 0.1457±0.0082  \\
\textit{DSAN} & 0.7567±0.0078 & 0.3840±0.0067 & 0.6089±0.0078  & 0.1363±0.0055 \\
\midrule
\textbf{Ours}\\
\textit{KISA-W/O-Know.} & 0.7518±0.0198 & 0.3812±0.0040 & 0.6142±0.0019 & 0.1436±0.0024 \\
\textit{KISA-Hour} & 0.7769±0.0071 & 0.4575±0.0293 & 0.6377±0.0131 & 0.1582±0.0057 \\
\textit{KISA-CardType} & 0.7761±0.0045 & 0.4535±0.0245 & 0.6371±0.0065 & 0.1557±0.0013  \\
\textit{KISA} & \textbf{0.7854${}^{*}$±0.0037} & \textbf{0.4808${}^{*}$±0.0013} & \textbf{0.6748${}^{*}$±0.0055} & \textbf{0.1645${}^{*}$±0.0025} \\
\bottomrule
\end{tabular}}
\vspace{-0.8em}
\label{table:Fraud Detection Task Result}
\end{table*}

\subsubsection{Results}
\label{sec:results}
We conduct experiments on the baselines and \textit{KISA}, the results are in Table \ref{table:Fraud Detection Task Result}. We divide these baselines into three parts: (1) without domain adaptation loss: \textit{LSTM} and \textit{LSTM (FT)}; (2) with global domain adaptation loss: \textit{MMD} and \textit{KL}; (3) with categorical subdomain adaptation loss: \textit{Transfer-HEN} and \textit{DSAN}. The experimental results offer us the following insightful observations: 

First, \textit{LSTM} surpasses \textit{LSTM (FT)}, which shows the `negative transfer' issue, indicating that the difference between the source and target domains is huge and directly transferring source model parameters may rapidly deteriorate the target models. 
Second, by conducting global domain adaptation, \textit{MMD} and \textit{KL} are better than \textit{LSTM}, showing that domain adaptation is more suitable than fine-tuning to alleviate the `negative transfer' issue. 
Third, \textit{Transfer-HEN} and \textit{DSAN} incorporate categorical information and get superior performance than \textit{MMD} in the setting of S1 $\rightarrow$ T1. However, they do not significantly outperform \textit{MMD} when transferring knowledge from S2 to T2, which is probably because T2 has fewer sequences than T1. It is difficult to construct the categorical subdomains well when lacking class labels. 
Moreover, \textit{KISA-Hour} and \textit{KISA-CardType} have a consistent improvement compared to the baselines by aligning the subdomains constructed by domain knowledge. Lastly, by integrating two domain knowledge, \textit{KISA} significantly ($p<0.05$) outperforms the best baseline, demonstrating its effectiveness.

\subsection{Cross-city Taxi Demand Prediction}
\label{spatiotemporal_task}
\subsubsection{Demand Prediction Dataset and Task}
The taxi demand datasets are from the DiDi GAIA open research collaboration project\footnote{\url{outreach.didichuxing.com}}, including the ride-sharing order data in Xi’an and Chengdu, China. There are about 6 and 8 million historical records from 2016.10 to 2016.11 for Xi’an and Chengdu respectively, containing taxi order messages including start location and start time. The location information is represented by longitude and latitude, and these location data cover the central city area of Xi’an and Chengdu. We divide the whole area into 16 $\times$ 16 grids as Zhang et al. \cite{zhang2017deep}, each grid has a size of 0.5km $\times$ 0.5km. 

\begin{table}[htbp]
    \caption{Statistics of the taxi demand datasets.}
    \vspace{-1em}
    \begin{tabular}{lcccc}
        \hline
           & Attributes & Xi'an & Chengdu      \\ \hline
         & Time span  &  2016.10-2016.11  & 2016.10-2016.11 \\
         & \# of records   & 5,922,961  &  8,439,537  \\
         & \# of stations & 16 $\times$ 16  & 16 $\times$ 16  \\
        \hline
    \end{tabular}
    \vspace{-.8em}
    \label{table: taxi demand}
\end{table}

Both two datasets have two months of historical records. The last 10\% duration in each dataset is test data, and the 10\% data before the test is for validation. The source domain utilizes all the rest data for training while the target domain only holds 1 or 3-day historical data for training. The task is to predict the taxi demand at the next hour for each grid. The statistics are listed in Table \ref{table: taxi demand}.

\subsubsection{Baseline Models}

As cross-city traffic prediction has many specialized state-of-the-art methods \cite{regionTrans2019,metaST_2019,crosstres_2022}, we mainly compare \textit{KISA} to these methods. 

\begin{itemize}[leftmargin=0.24cm]

\item \textbf{\textit{ARIMA}} \cite{arima_2003} is a widely used time series prediction model, considering the demand observations of the recent 24 slots.

\item \textbf{\textit{GBRT}} \cite{li_bike_2015} and \textbf{\textit{XGBoost}} \cite{xgboost_chen_2016} are tree-based learners that take the same input features as \textit{ARIMA}.

\item \textbf{\textit{LSTM}} \cite{ma_2015_LSTM} feeds the demand observations of the recent 24 slots into the LSTM network and gets predictions by two MLP layers.

\item \textbf{\textit{STMeta}} and \textbf{\textit{STMeta (FT)}} \cite{wang2021exploring} are spatiotemporal prediction models, considering both temporal and spatial factors. \textit{STMeta} trains by target domain data while \textit{STMeta (FT)} trains based on source domain data and then fine-tunes in the target domain. 

\item \textbf{\textit{RegionTrans}} \cite{regionTrans2019} is a cross-city transfer learning method that enables region-level knowledge transfer. It constructs the subdomain by intuitive division (i.e., the same geographic location).

\item \textbf{\textit{MetaST}} \cite{metaST_2019} is a transfer learning method that utilizes meta-learning for source training and transfers the region-level spatiotemporal knowledge from source cities. 

\item \textbf{\textit{CrossTRes}} \cite{crosstres_2022} 
 transfers region-level knowledge by re-weighting source regions. For a fair comparison, we adopt proximity and function graph as \textit{STMeta} to learn regional spatial embedding.
\end{itemize}

\subsubsection{Evaluation Metric} We exploit the widely used metric, namely RMSE (Root Mean Square Error) to assess the performance of prediction models \cite{zhang2017deep,regionTrans2019,metaST_2019}.

\subsubsection{Implementation Details} 
For fair comparisons, all baseline methods use the same input (i.e., the demand observations of the recent 24 hourly slots). 
The backbone network structure for \textit{KISA} is \textit{STMeta}.
For the considerations of spatial knowledge, we build two kinds of graphs as Wang et al. \cite{wang2021exploring} (i.e., proximity and function graph). The proximity graphs are calculated based on the Euclidean distance. The function graphs are computed by the Pearson coefficient of the time series of stations. The hidden states of the \textit{STMeta} network are 64 (the dimension of spatiotemporal representations). The degree of graph Laplacian is 1. We use Adam \cite{adam_2015} as the optimizer to train the network. The learning rate and batch size are set to $10^{-5}$ and 32 respectively.

\subsubsection{Domain-knowledge Exploration \& Exploitation} \textit{KISA} leverages the following knowledge for the demand prediction task:

\textbf{\textit{Knowledge 1: Spatial Proximity}}. As the ‘First Law of Geography’ says, ‘\textit{Everything is related to everything else. But near things are more related than distant things}’. Proximity has been extensively used in spatiotemporal prediction tasks \cite{graph_wavenet_2019, Song_Lin_Guo_Wan_2020}. To utilize this knowledge, in \textit{KISA-S.P.}, the knowledge-based features are the row and columns index of grids and nearby grids are encouraged to group into the same subdomain. We construct 40 subdomains in both source and target domains.

\textbf{\textit{Knowledge 2: Temporal Heterogeneity}}. The transportation demand can vary greatly for the same station at different periods (heterogeneous temporal patterns) \cite{zhang2017deep,Guo_Lin_Feng_Song_Wan_2019,wang2021exploring}. In \textit{KISA-T.H.}, subdomains contain grids with similar daily demand patterns and we use the average daily demand as the knowledge-based feature. For our 60-minute demand prediction task, we extract features with 24 dimensions, where each dimension corresponds to one hour within a day. We built 32 subdomains in both source and target domains.

\subsubsection{Results}
Table~\ref{table:demand_prediction_results} shows the results. We observe that deep learning models (\textit{LSTM} and \textit{STMeta}) suffer from the data scarcity issue in the target domain and perform poorly, and \textit{XGBoost} show its learning capability at small data scenarios. 
When the data for the target city grow and the domain-specific pattern becomes more pronounced, directly transferring the model (i.e., \textit{STMeta (FT)}) may encounter the negative transfer issue (e.g., 3-day results from Chengdu to Xi'an). 
Meanwhile, \textit{RegionTrans} gets remarkable results and shows that transferring region-level knowledge help alleviate the negative transfer issue. 
\textit{KISA-S.P.} is better than \textit{RegionTrans}, which demonstrates the superiority of subdomain-level transfer. 
\textit{MetaST}, a multi-source transfer algorithm, does not perform well in our tasks perhaps due to the usage of only one source domain. Besides, \textit{CrossTReS} reweights source regions for transfer which effectively alleviates the `negative transfer' issue and performs better than all other baselines. 
More importantly, \textit{KISA-T.H.}, by transferring knowledge from subdomains with similar temporal patterns, achieves lower error compared to all baselines. Moreover, by incorporating more domain knowledge, \textit{KISA} consistently outperforms the best baseline, where the largest improvement is reducing RMSE by up to 3.12\%, verifying its generalizability and effectiveness.

\begin{table}[htbp]
    \caption{Results (RMSE) on the demand prediction task. The target city has 1 or 3-day data. The best results are in bold.}
    \vspace{-0.8em}
\begin{tabular}{lcccccccccccccccccccc}
\toprule
 & \multicolumn{2}{l}{\textbf{Chengdu} $\rightarrow$ \textbf{Xi'an}} & \multicolumn{2}{c}{\textbf{Xi'an} $\rightarrow$ \textbf{Chengdu}} \\ 
 \cmidrule(lr){2-3}  \cmidrule(lr){4-5}
 & 1-day & 3-day & 1-day & 3-day \\ 
\midrule
\multicolumn{2}{l}{\textbf{Target Only}} \\
\textit{ARIMA} & 11.641 & 9.977 & 16.038 & 13.565 \\
\textit{GBRT} & 12.284 & 10.353 & 13.405 & 12.039 \\
\textit{XGBoost} & 11.452 & 9.665 & 11.496 & 10.985 \\
\textit{LSTM} & 12.382 & 10.451 & 14.112 & 10.013 \\
\textit{STMeta}  & 14.232 & 8.444 & 15.337 & 10.167 \\
\midrule
\multicolumn{2}{l}{\textbf{Source \& Target}} \\
\textit{STMeta (FT)}   & 11.226 & 9.225 & 10.020 & 9.890 \\
\textit{RegionTrans} & 10.448 & 8.195 & 9.648 & 9.393 \\
\textit{MetaST}  & 10.882 & 8.500 & 10.648 & 9.966 \\
\textit{CrossTReS}  & 10.235 & 8.281 & 9.518 & 9.384 \\
\midrule
\multicolumn{2}{l}{\textbf{Ours}} \\
\textit{KISA-S.P.}  & 10.395 & 8.151 & 9.615 & 9.356 \\
\textit{KISA-T.H.}  & 10.013 & 8.114 & 9.437 & 9.253 \\
\textit{KISA}  & \textbf{9.916} & \textbf{8.054} & \textbf{9.391} & \textbf{9.186} \\
\bottomrule 
\end{tabular}
\vspace{-1.4em}
\label{table:demand_prediction_results}
\end{table}

\subsection{Analysis}
\subsubsection{Comparison of Domain Knowledge Utilizing}
The key improvement of \textit{KISA} in utilizing domain knowledge is constructing knowledge-inspired subdomains for domain adaptation. 
To analyze the effectiveness of knowledge-inspired subdomains, we implement a variant called \textbf{\textit{KISA-W/O-Know}.} (\textit{KISA} without Knowledge), which constructs categorical subdomains (in fraud detection, just two subdomains including positive or negative samples) rather than by knowledge. This variant applies the same distance discrepancy and alignment loss with \textit{KISA}. 
In Table~\ref{table:Fraud Detection Task Result}, we observe that \textit{KISA-W/O-Know.} rapidly deteriorates and can only achieve performance commensurate with baselines. This verifies the importance of using domain knowledge to construct fine-grained subdomains for knowledge transfer.
On the other hand, compared to \textit{KISA-CardType} and \textit{KISA-Hour}, \textit{KISA} fully utilizes two domain knowledge factors and achieves better performance. This inspires future research to explore more domain knowledge for better performance.

\subsubsection{Latent Feature Visualization} 
\begin{figure}[htbp]
    \centering
    \subfigure[\textit{MMD}]{
        \includegraphics[width=0.152\textwidth]{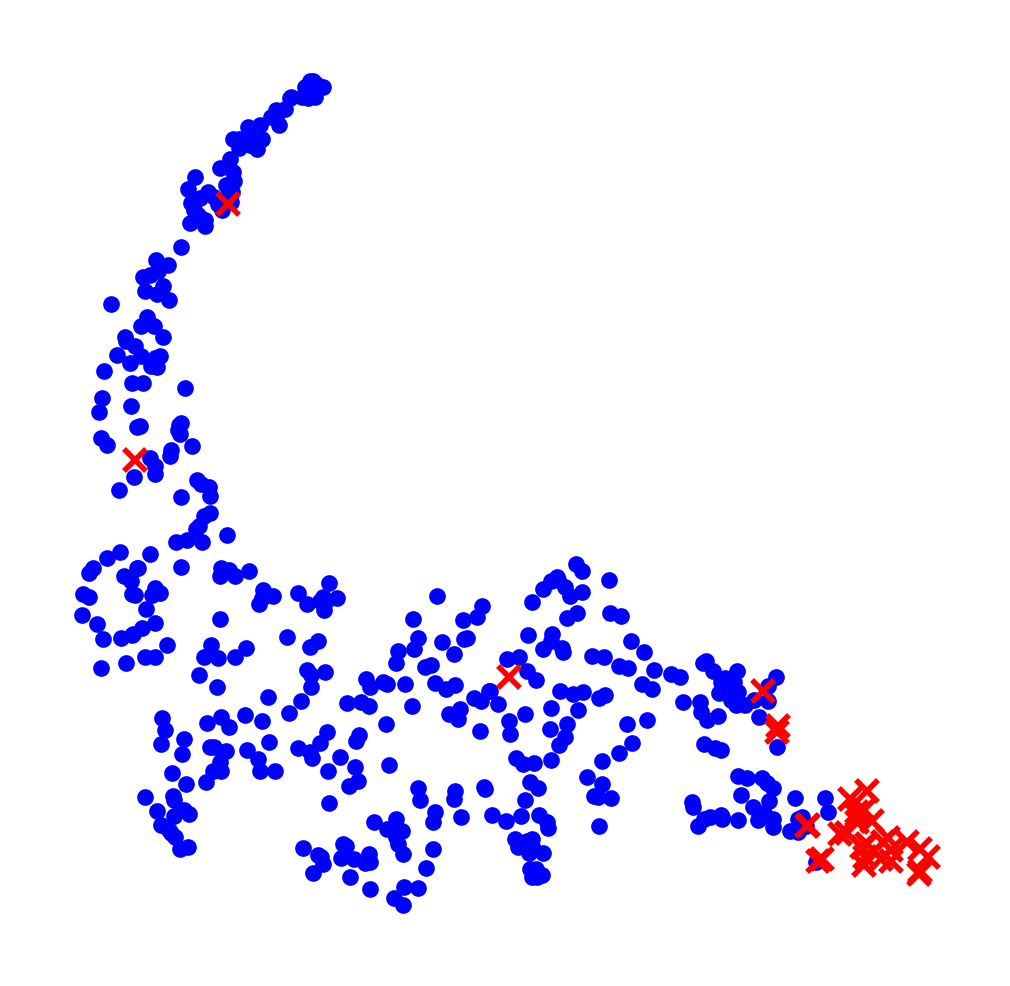}
        \label{vis_mmd}
    }\hspace{-2mm}
    \subfigure[\textit{DSAN}]{
	\includegraphics[width=0.152\textwidth]{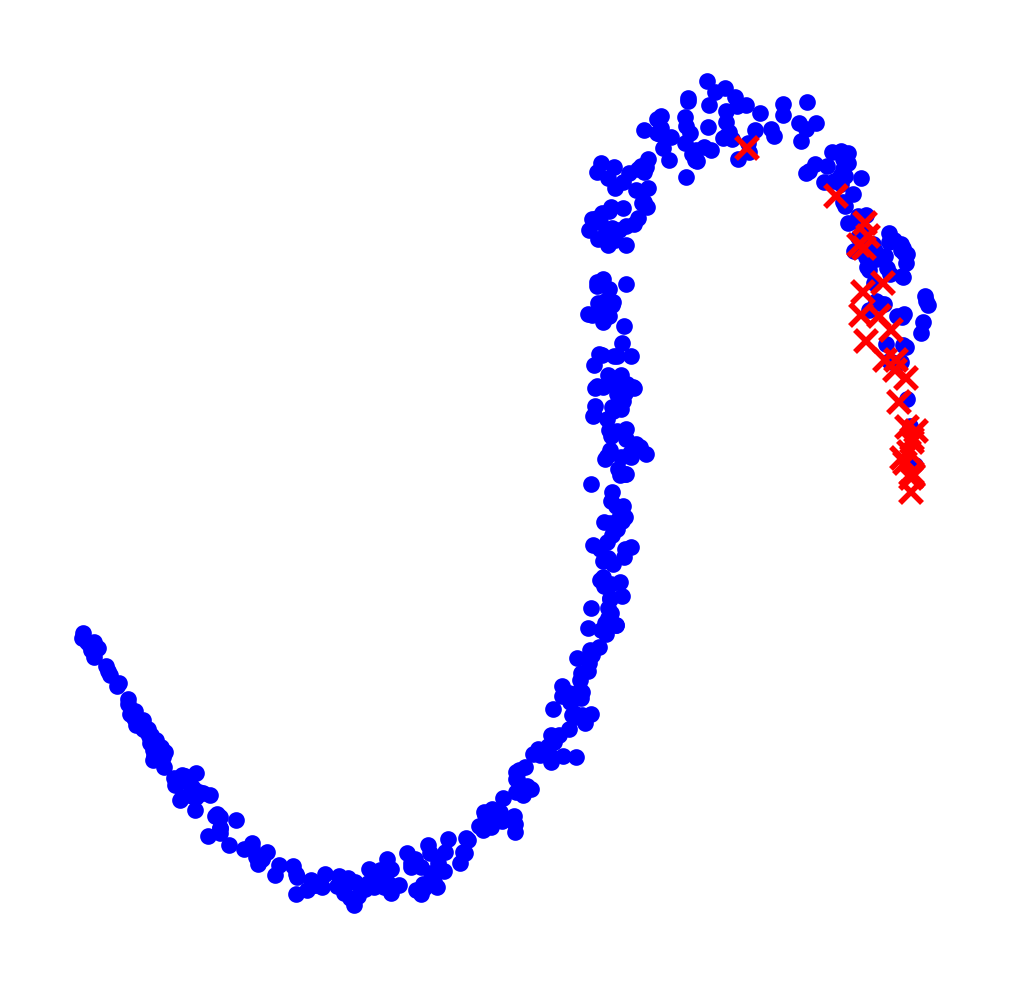}
        \label{vis_dsan}
    }\hspace{-2mm}
    \subfigure[\textit{KISA}]{
        \includegraphics[width=0.152\textwidth]{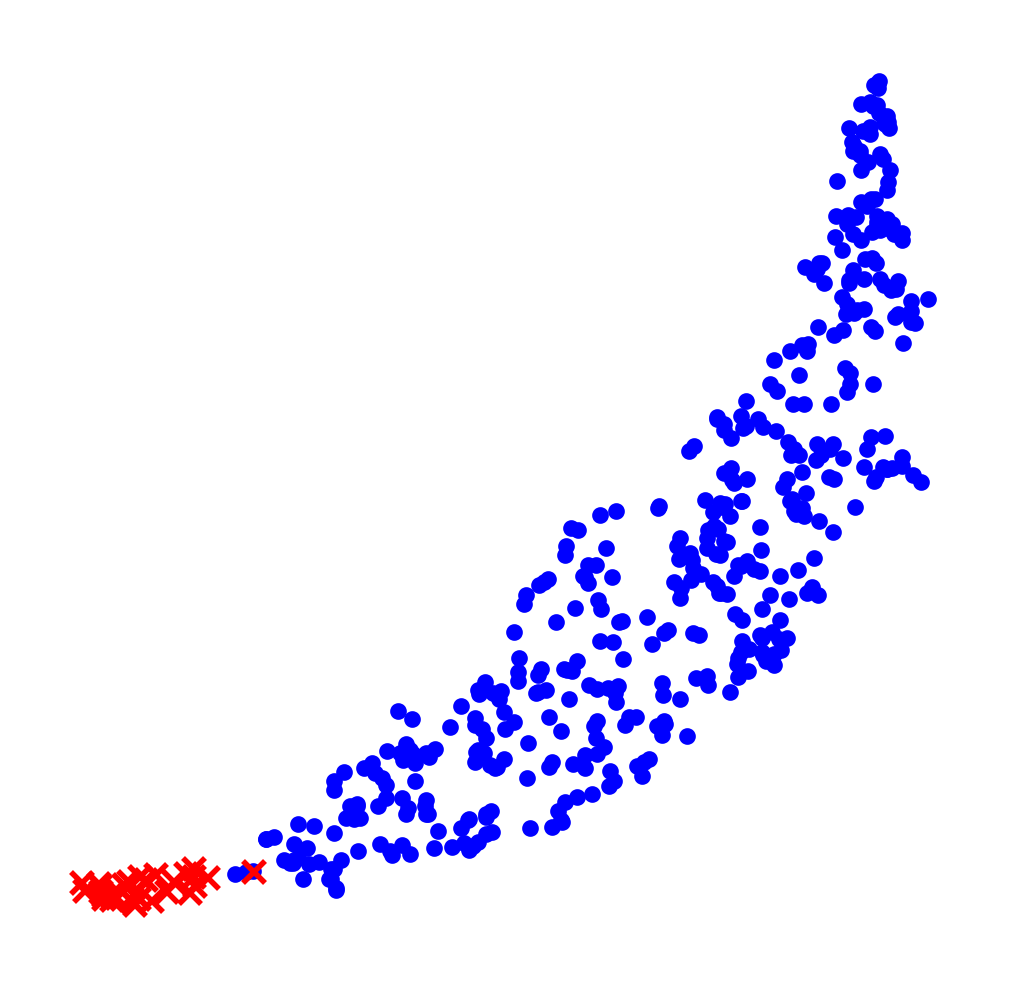}
        \label{vis_kisa}
    }
    \vspace{-1em}
    \caption{Latent feature visualization using \textit{t-SNE} for \textit{MMD}, \textit{DSAN}, and \textit{KISA}. The visualization is conducted on fraud detection tasks under the setting of S1 $\rightarrow$ S2. Different colors indicate different classes (red cross: fraud, blue round: normal). Best viewed in color.}
    \vspace{-1.3em}
    \label{vis}
\end{figure}

Based on the fraud detection task (S1$\to$T1), we present a visualization of the latent representations obtained from three domain adaptation methods, namely \textit{MMD} \cite{Tzeng_invariance}, \textit{DSAN} \cite{zhu2020deepsubdomain}, and \textit{KISA}.
We choose these three methods to study because they employ the same domain divergence measure (i.e., MMD). \textit{t-SNE} technique \cite{t_sne_2014} is adopted to project high-dimension representation into 2-D space. 
In Fig. \ref{vis}, the red cross and blue round represent fraud and normal samples, respectively. 
Fig. \ref{vis_mmd} shows the result for \textit{MMD}, which aligns the global distribution across two domains. It shows that several fraud samples are entangled with normal samples and thus are hard to classify. Fig. \ref{vis_dsan} shows the result for \textit{DSAN}, a categorical subdomain adaptation method. It effectively aggregates the class manifolds, yet a substantial overlap persists between the domains of fraud and normal samples. In contrast, Fig. \ref{vis_kisa} presents the learned representations by \textit{KISA}, wherein a distinct boundary between fraud and normal samples is discernible. Remarkably, \textit{KISA} successfully compresses the fraud manifold while minimizing the mixing of fraud samples with normal ones. Consequently, these findings suggest that \textit{KISA} exhibits superior capabilities in learning more robust and distinguishable representations compared to \textit{MMD} and \textit{DSAN}.

\begin{figure*}[t]
\begin{minipage}[c]{0.685\linewidth}
\centering
\includegraphics[width=1\linewidth]{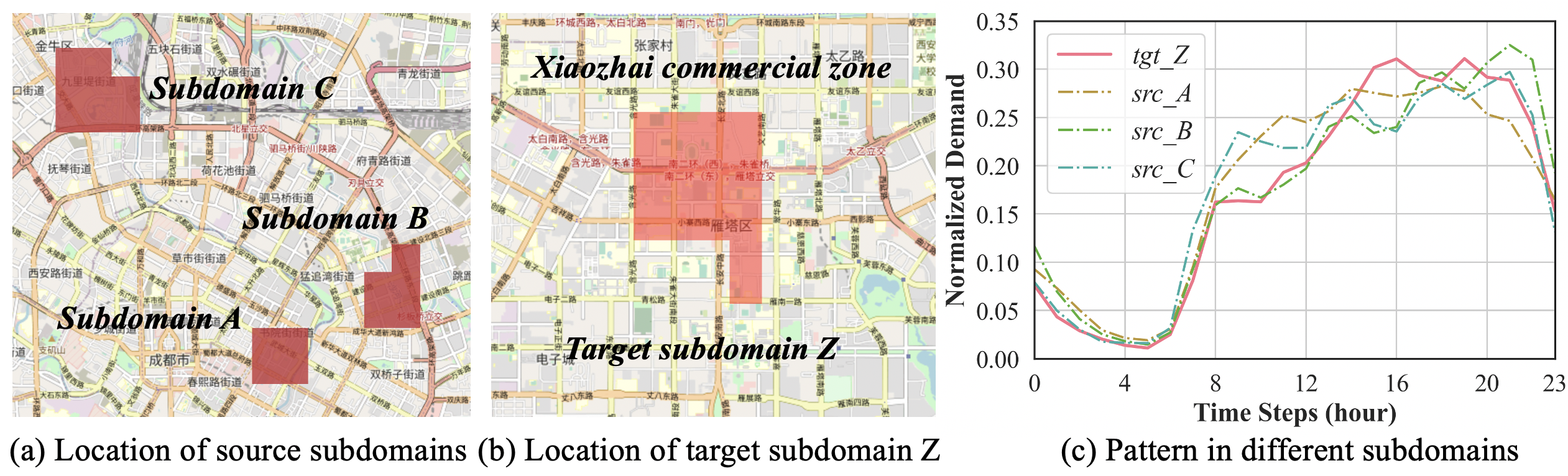}
\vspace{-1.9em}
\caption{Case study of subdomain matching under the setting of Chengdu $\rightarrow$ Xi'an.}
\label{fig: case_study}
\vspace{-0.7em}
\end{minipage}
\hspace{0.5mm}
\begin{minipage}[c]{0.27\linewidth}
\includegraphics[width=1\linewidth]{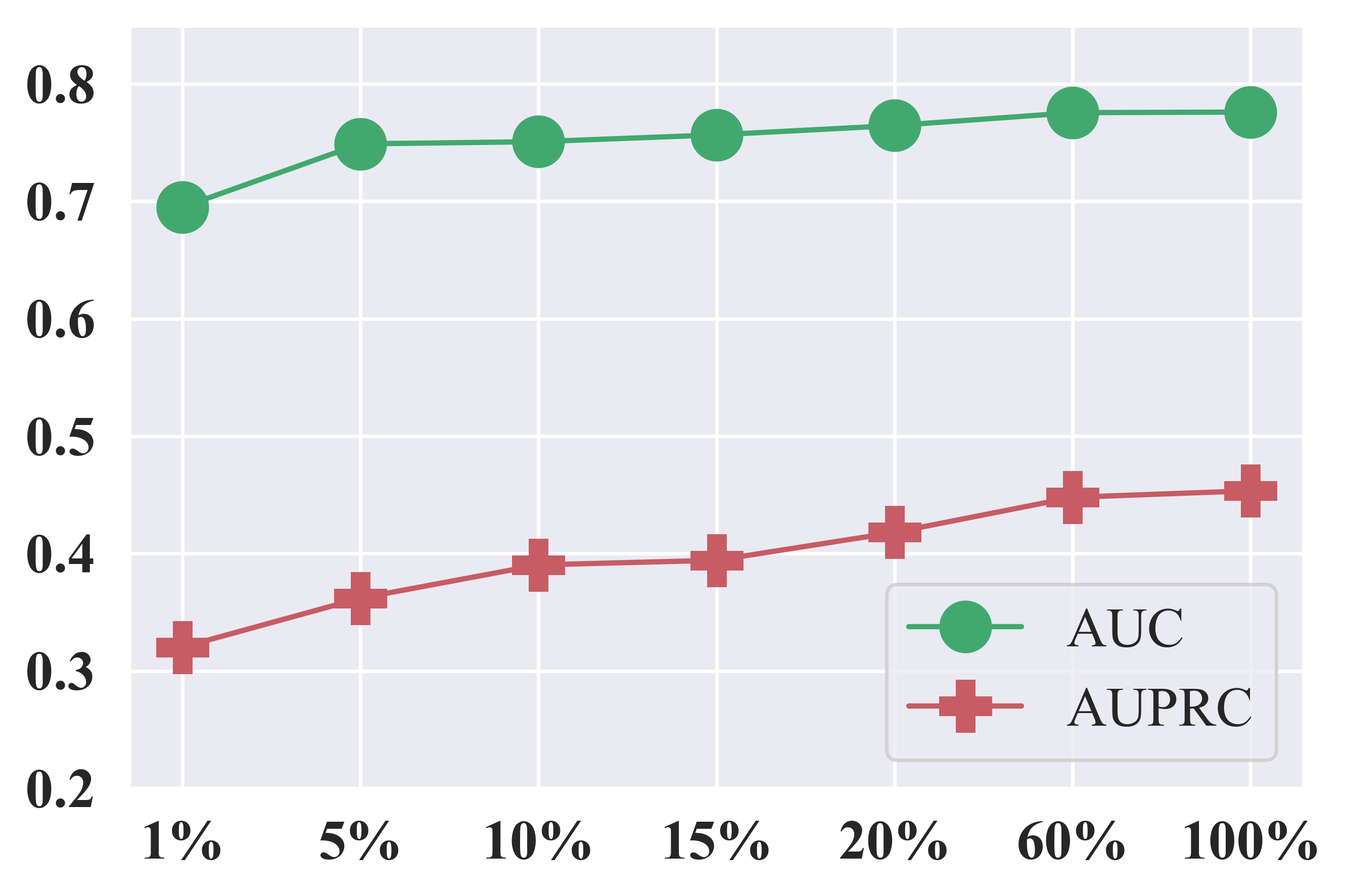}
\caption{Performance vs. source training data size.}
\label{fig: scala_performance}
\end{minipage}
\vspace{-.7em}
\end{figure*}

\subsubsection{Case Study of Subdomain Matching} 
To investigate the generated subdomains and their associated matching relationships, we conduct a case study in the demand prediction task under the setting of Chengdu $\rightarrow$ Xi'an. 
We chose this task because it allows for easy visualization of subdomains (i.e., clustering of spatial grids) in figures. The subdomains used in this study are from \textit{KISA-S.P.}. Fig.~\ref{fig: case_study} shows the geographical distribution of target subdomain Z and its corresponding source subdomains A, B, and C.
We observe that each subdomain consists of adjacent grids (spatial proximity). Fig.~\ref{fig: case_study} depicts the target subdomain Z in Xi'an, which includes the Xiaozhai commercial zone where mall openings result in a surge of taxi demand after 10 a.m., peaking in the late afternoon. Additionally, daily patterns of source subdomains A, B, and C in Chengdu exhibit remarkable similarities.
This observed similarity between source subdomains and their target counterparts demonstrates the effectiveness of \textit{KISA} in identifying source subdomains that enable focused, fine-grained transfer learning. By aligning subdomains selectively, \textit{KISA} reduces the risk of "negative transfer" problems that can arise from global domain adaptation approaches.

\subsubsection{Sensitivity on Training Data Size}
We train different models\footnote{We choose \textit{KISA-CardType} in the setting of S1 $\rightarrow$ T1 to conduct this experiment.} by varying the training data size of the source domain. In particular, we sample 1\%, 5\%, 10\%, 15\%, 20\%, 60\%, and 100\% from the training data and collect the associated AUC and AUPRC of the test data in the target domain. From Fig. \ref{fig: scala_performance}, we observe that the more training data we use, the better performance we get. Besides, our method is capable of transferring reasonable knowledge from much less source domain data. In our scenarios, we could get competitive performance with 20\% data (i.e., about 38.5k transaction records).

\section{Related Work}
\textbf{\textit{Global Domain Adaptation}}. 
There have been extensive efforts on \textit{global domain adaptation} during the past several years. The latest advances embed domain adaptation modules in deep feature learning networks to extract domain-invariant representations~\cite{ghifary2014domain,tzeng2015simultaneous,ganin2016domain,ganin2015unsupervised,pmlr_v37_long15,yan2017mind,shu2018dirt,long2017conditional}. 
Following the taxonomy from Zhu et al.~\cite{zhu2020deepsubdomain}, there are two main approaches: (i) \textit{statistic moment matching based approaches} conduct alignment according to the statistic distance between source and target domain (e.g., maximum mean discrepancy \cite{pmlr_v37_long15,long2017deep}, central moment discrepancy \cite{zellinger2017central}, and second-order statistics \cite{Sun2016DeepCC}); (ii) \textit{adversarial approaches} \cite{ganin2016domain,hoffman2018cycada} integrate two adversarial players similarly to Generative Adversarial Networks (GANs) \cite{gan_nips_2014}. A domain discriminator is learned by minimizing the classification error of distinguishing the source from the target domains, while a deep classification model learns transferable representations that are indistinguishable by the domain discriminator \cite{long2017conditional}. 
Compared to previous global domain adaptation methods, \textit{KISA} is one of the pioneering efforts toward fine-grained domain adaptation by incorporating domain knowledge.

\textbf{\textit{Subdomain Adaptation}}. 
Recently, there has been substantial interest and efforts \cite{long2017conditional,pei2018multi,kumar2018co,xie2018learning,regionTrans2019} for \textit{subdomain adaptation} which focuses on aligning the distributions of the relevant subdomains. 
According to the definition of subdomains, there are two main approaches: (i) \textit{class label based methods}: consider a subdomain as a set with the same label and most subdomain methods follow this paradigm. For example, by matching labeled source centroids and pseudo-labeled target centroids, \textit{MSTN} \cite{xie2018learning} learns semantic representations for unlabeled target samples. \textit{Co-DA} \cite{kumar2018co} builds several different feature spaces and aligns the source and target distributions in each of them separately while promoting alignments that concur with one another concerning the class predictions on the unlabeled target data. (ii) \textit{intuitive division approaches}: construct subdomains by domain-specific intuitions. 
For example, \textit{RegionTrans} \cite{regionTrans2019} treats intuitively divided fixed grids as subdomains and then aligns learned representation with the region-matching function using crowd flow and check-in data. 
Compared to previous subdomain adaptation methods, \textit{KISA} is a data-driven solution that can adaptively construct subdomains by providing the selected domain knowledge and samples, which extends the ability to perform fine-grained domain adaptation.

\section{Conclusion}
In this paper, we propose a novel transfer learning framework called \textit{KISA} by leveraging domain knowledge to enable fine-grained subdomain adaptation. 
We propose the knowledge-inspired subdomain division problem to construct subdomains and corresponding solutions to solve it. Moreover, \textit{KISA} introduces a knowledge fusion network to fully exploit diverse domain knowledge, which is more applicable in real-world applications. 
Finally, we prove the effectiveness of \textit{KISA} by conducting extensive experiments on fraud detection and demand prediction tasks.

\textbf{\textit{Limitations and future work}}.
The knowledge-inspired subdomain division problem takes in the predefined number of subdomains, which is given based on our experience. Whether there exists an optimal number of subdomains is still not yet known. In future work, we will try our best to give a theoretical analysis of how the subdomain number affects the performance of subdomain adaptation and test \textit{KISA}'s generalizability for more applications.

\begin{acks}
This work was supported by National Science Foundation of China (NSFC) Grant No. 61972008 and Ant Group.
\end{acks}

\bibliographystyle{ACM-Reference-Format}
\balance
\bibliography{ref}

\end{document}